\documentclass{article}

\usepackage{arxiv}

\usepackage[utf8]{inputenc} % allow utf-8 input
\usepackage[T1]{fontenc}    % use 8-bit T1 fonts
\usepackage{hyperref}       % hyperlinks
\usepackage{url}            % simple URL typesetting
\usepackage{booktabs}       % professional-quality tables
\usepackage{amsfonts}       % blackboard math symbols
\usepackage{nicefrac}       % compact symbols for 1/2, etc.
\usepackage{microtype}      % microtypography
\usepackage{lipsum}		% Can be removed after putting your text content
\usepackage{graphicx}
\usepackage{doi}
\usepackage{amsmath}
\usepackage{placeins}
\usepackage{tikz}
\usetikzlibrary{positioning, decorations.pathreplacing, calligraphy, calc}

\author{Bouarfa Mahi \\ Quantiota \\ Email: info@quantiota.org}

\title{Structured Knowledge Accumulation: The Principle of Entropic Least Action in Forward-Only Neural Learning}

\author{ \href{https://orcid.org/0009-0008-7158-2729}{\includegraphics[scale=0.06]{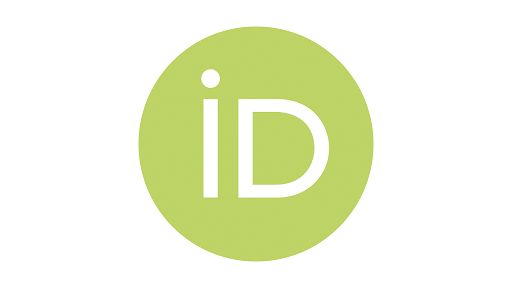}}\hspace{1mm}Bouarfa Mahi Quantiota \thanks{Use footnote for providing further
		information about author (webpage, alternative
		address)---\emph{not} for acknowledging funding agencies.} \\
	Université Joseph Fourier\\
Grenoble, Auvergne-Rhône-Alpes, FR\\
	\texttt{info@quantiota.org} }
\renewcommand{\shorttitle}{SKA: The Principle of Entropic
Least Action in Forward-Only Neural Learning}

\hypersetup{
pdftitle={A template for the arxiv style},
pdfsubject={q-bio.NC, q-bio.QM},
pdfauthor={David S.~Hippocampus, Elias D.~Striatum},
pdfkeywords={First keyword, Second keyword, More},
}

\begin{document}
\maketitle

\begin{abstract}
This paper aims to extend the Structured Knowledge Accumulation (SKA) framework recently proposed by \cite{mahi2025ska}. We introduce two core concepts: the Tensor Net function and the characteristic time property of neural learning. First, we reinterpret the learning rate as a time step in a continuous system. This transforms neural learning from discrete optimization into continuous-time evolution. We show that learning dynamics remain consistent when the product of learning rate and iteration steps stays constant. This reveals a time-invariant behavior and identifies an intrinsic timescale of the network. Second, we define the Tensor Net function as a measure that captures the relationship between decision probabilities, entropy gradients, and knowledge change. Additionally, we define its zero-crossing as the equilibrium state between decision probabilities and entropy gradients. We show that the convergence of entropy and knowledge flow provides a natural stopping condition, replacing arbitrary thresholds with an information-theoretic criterion. We also establish that SKA dynamics satisfy a variational principle based on the Euler-Lagrange equation. These findings extend SKA into a continuous and self-organizing learning model. The framework links computational learning with physical systems that evolve by natural laws. By understanding learning as a time-based process, we open new directions for
building efficient, robust, and biologically-inspired AI systems.
\end{abstract}

% keywords can be removed
\keywords{Structured Knowledge Accumulation \and Time-Invariant Learning \and Tensor Net Function \and Continuous-Time Neural Dynamics \and Forward-Only Learning}

\section{Introduction and Preliminaries}
Understanding the fundamental principles that govern how artificial neural networks accumulate knowledge is a cornerstone of modern machine learning research. While most learning algorithms are framed as discrete optimization routines, this perspective often overlooks the deeper, time-evolving structure underlying knowledge formation in neural systems. Bridging this gap requires a paradigm that not only captures the flow of information through a network but also aligns with the natural principles of dynamical systems. The SKA framework offers such a perspective, reimagining learning as a continuous, forward-only process governed by entropy gradients and intrinsic timescales. This approach not only challenges conventional backpropagation-based paradigms but also provides a more physically-grounded interpretation of learning, opening new avenues for theory, architecture design, and hardware implementation.\\
\indent The SKA framework, introduced in our previous work \cite{mahi2025ska}, redefines entropy as a dynamic, layer-wise measure of knowledge alignment in neural networks. By formulating entropy as 
$$H^{(l)} = -\frac{1}{\ln 2} \sum_{k} \mathbf{z}^{(l)}_k \cdot \Delta\mathbf{D}^{(l)}_k,$$ 
SKA provides an alternative to traditional backpropagation-based learning through forward-only, local entropy minimization.
\indent While our initial work established the theoretical foundation of SKA, demonstrating how the sigmoid function emerges naturally from continuous entropy minimization and how learning can proceed without backpropagation, several profound aspects of the framework remained unexplored. \\
This article presents two significant advancements in our understanding of SKA dynamics: First, we reinterpret the learning rate parameter $\eta$ as a time step $\Delta t$ in a continuous dynamical system. This perspective transforms our understanding of neural learning from discrete optimization steps to continuous temporal evolution. We present compelling empirical evidence that SKA exhibits time-invariant behavior when the product $\eta \times K$ (time step $\times$ number of steps) remains constant, revealing an intrinsic characteristic timescale of the network. 
Second, we define the Tensor Net function, which is a new metric that describes the relationship among the decision probabilities, knowledge evolution, and  entropy. This function establishes a principled criterion for deciding when structured knowledge accumulation starts, with its zero-crossing point showing a transition to the organized learning. In contrast, the convergence of entropy and knowledge flow indicates the point when learning is finished, providing an emergent stopping condition which is based on the information-theoretic equilibrium instead of arbitrary thresholds. \\
\indent These findings extend SKA beyond a mere alternative training technique to a framework that connects physical systems governed by natural laws with computational neural networks. As we will demonstrate, these phenomena emerge naturally from a principle of entropic least action. By comprehending neural learning as a continuous-time procedure with intrinsic dynamics, we provide new research directions for designing more efficient, biologically plausible AI systems.

\subsection{Core Definitions in the SKA Framework}
Before exploring the time-continuous formulation, we clarify the foundational concepts of the
SKA framework to ensure a consistent interpretation throughout the paper. While these terms are already discussed in the literature, we introduce additional precision to align them with the specific context and requirements of the SKA framework:
	
	\begin{itemize}
		\item \textbf{Knowledge ($\mathbf{z}$)} refers to the pre-activation values in each neuron that serve as inputs to the sigmoid function, mathematically represented as tensor $\mathbf{Z}$ or vectors $\mathbf{z}^{(l)}$ for each layer $l$. This knowledge accumulates over forward passes and directly influences decision probabilities through sigmoid transformation.
		
		\item \textbf{Structured Knowledge} describes knowledge ($\mathbf{z}$) that has aligned with decision probability shifts ($\Delta\mathbf{D}$) in a way that minimizes layer-wise entropy. It is characterized by specific relationships between knowledge vectors and entropy gradients, and becomes identifiable when the Tensor Net function crosses zero, indicating $\int \mathbf{D}^{(l)} \, dz = \mathbf{H}^{(l)}$.
		
		\item \textbf{Structured Knowledge Accumulation} is the process by which knowledge progressively increases in a structured manner, leading to enhanced decision probabilities that better discriminate between classes. This accumulation manifests as growth in knowledge magnitude (measured by the Frobenius norm of $\mathbf{Z}$) that correlates with increased separation in output decision probabilities. The process follows time-invariant trajectories with characteristic timescales and proceeds through forward-only learning, without requiring backpropagation.
		\end{itemize}
	
	These mathematically precise definitions distinguish SKA's conceptual framework from traditional learning paradigms and provide the foundation for our subsequent analysis of time-invariant properties and equilibrium dynamics.

\subsection{Literature Review}

The basic concepts of neural learning have mainly revolved around optimization methods utilizing the gradient and backpropagation systems \cite{rumelhart1986learning}. While these methods have provided important progress in the field of deep learning, they do not help explain the processes of knowledge accumulation and the flow of learning's dynamic quite effectively. Because of this gap in research, some studies have attempted to approach the
problem by framing learning as a physically-inspired process or one rooted in nature.  

In this direction, perhaps the most advanced idea is the notion of forward-only learning strategies, which aim to both increase biological realism and simplify computation \cite{lillicrap2016random, terres2024ecai}. The F-OAL (Forward-only Online Analytic Learning) model, for example, has been shown to employ resources very economically and train quickly in incremental environments \cite{zhuang2024f}. In the same way, the performance of fully-spiking neural networks under forward-only learning has been studied in open-world conditions \cite{terres2024}, where more recent work has sought to apply neural polarization methods to enhance instabilities and improve generalization within these systems \cite{terres2024ecai}.

Understanding the problem of learning as a continuous-time dynamical system is yet another shift in the literature. Neural ODEs \cite{chen2018neural} are related to formal neural reasoning using equations that compute with neural networks by embedding differentiation and thus ‘synchronizing’ the evolution of the network to the dynamics of the system. Yang \cite{yang2019physical} built on this by looking at the scaling limits of wide neural networks and provided a theoretical justification based on behaviors of a Gaussian process and neural tangent kernels. In a closely related approach, Behrmann et al. \cite{behrmann2019invertible} studied invertible residual networks where transformations are information preserving, allowing for the free transport of information in deep systems.

Alternatives to optimization have included the exploration of entropy and energy-based learning models. Balduzzi \cite{balduzzi2017mechanistic} along with LeCun et al. \cite{lecun2006tutorial} pointed out that thermodynamics and energy principles can constructively add on the evolution of learning, while Scellier and Bengio \cite{scellier2017equilibrium} proposed equilibrium propagation – linking energy-based models and backpropagation to do so. These models align with the Structured Knowledge Accumulation (SKA) framework that views the neural learning rationale as entropy minimization \cite{mahi2025ska}.

Considering feedback-based motor learning models from a biological standpoint, it has been illustrated how neural networks might apply organized changes through real-time feedback \cite{feulner2025}. The contribution of the dendrites in the refinement and optimization of learning processes in artificial neural networks has also been highlighted \cite{chavlis2025}. Research conducted on neural manifolds indicates that sophisticated behavior can arise from low-dimensional configurations within the network \cite{gallego2017neural}.

Knowledge enhancement through symbol neural hybrids and knowledge distillation still maintains their importance towards the interpretability and scalability on a neural learning system. Shavlik \cite{shavlik1994} looked at the problem of combining symbolic rules with a neural nets and Shen et al. \cite{shen2025} suggested a distillation based meta learning strategy for graph neural networks. Zhuang et al. \cite{zhuang2025} also proposed a new analytic formulation of learning pattern recognition in convolutional neural networks.

Cognitive science frameworks combine history with computational models, which provides unique value. Delahunt et al. \cite{2} designed bio-inspired models around the learning of moth olfaction, whereas some earlier dissertation research focused on the fundamentals of the theory of artificial neural networks \cite{3}. In \cite{1}, Ye et al. describe SequenceMorph, which is an unsupervised learning technique aimed at motion tracking in medical imaging.

Together, these contributions collectively highlight the need for a unifying framework that can synthesize forward-only dynamics, entropy-based learning, and biologically plausible processes. The SKA framework aims to address this gap by offering a structured, time-continuous model of learning that bridges computational and physical systems \cite{mahi2025ska}.
\subsection{Research Gap and Motivation}
While the existing body of literature provides significant insights into continuous-time learning, entropy-driven dynamics, and biologically plausible forward-only architectures, a unified framework that systematically integrates these perspectives remains underdeveloped. Most existing models either lack a principled metric of structured knowledge accumulation or fail to establish time-invariant properties that govern the evolution of learning as a natural physical process. Furthermore, despite advances in differential formulations of neural dynamics, a comprehensive approach that captures the onset, progression, and convergence of knowledge flow within a self-referential, entropy-regulated learning system has yet to be formalized. This research addresses this gap by proposing the SKA framework, which introduces new mathematical tools and properties—such as the Tensor Net function and the characteristic time of learning—that offer a foundational shift in understanding neural learning as a continuous-time dynamical system governed by natural laws.

This paper is organized as follows: After the first section of introduction and preliminaries, the subsection~1.2 provides a detailed literature review, mentioning prior work related to entropy-based learning, continuous-time neural dynamics, biologically inspired computational models, and forward-only learning mechanisms. The subsection~1.3, titled research gap and motivation, outlines the limitations in existing approaches and establishes the rationale behind the proposed SKA framework. The remainder of the paper is organized as follows: Section~2 presents the core findings and theoretical contributions of this study, including the reinterpretation of the learning rate as a time step, the introduction of characteristic time as an intrinsic learning property, a formulation of knowledge flow dynamics, and the definition of the Tensor Net function as a criterion for structured learning onset. This section also discusses variational analysis and the practical implications and broader applications of the proposed framework. Finally, Section~3 concludes the paper by summarizing the main contributions and outlining future directions for advancing the SKA framework and its integration into more complex learning systems.

\section{Our Obtained Results}
This section presents the core findings of our study. It is organized into five subsections that explore key conceptual and analytical developments, including a reinterpretation of the learning rate, the formulation of characteristic time, the modeling of knowledge flow, the introduction of the tensor net function, and relevant implications and applications.
\subsection{Reinterpreting Learning Rate as Time Step}

\subsubsection{From Learning Rate to Time Step}

One of the insights we have found so far on SKA is that the learning rate parameter \( \eta \) can be seen as a time step \( \Delta t \) in a continuously implemented dynamical system. Instead of considering learning to be a sequence of optimization steps, this approach sees learning as a smooth temporal evolution that is controlled by the system’s dynamics. In traditional neural networks, \( \eta \) is typically viewed as an arbitrary hyperparameter controlling step size in gradient
descent optimization. Still, within the context of the SKA framework, we can identify how entropy minimization follows a specific progression. This indicates that \( \eta \) plays a more profound role, such as that of the discretization step in a much more underlying continuous time process.
	
To formalize this interpretation, we begin by considering the entropy-driven weight update in SKA. Instead of defining learning as an iterative parameter adjustment, we express it in the form of a time-dependent evolution equation:
	
	\begin{equation}
		\frac{\Delta \mathbf{W}^{(l)}}{\Delta t} = -\nabla_{\mathbf{w}} H^{(l)}.
	\end{equation}
	
	This equation resembles a continuous-time dynamical system, where the rate of weight change is proportional to the negative gradient of entropy. In the limit as \( \Delta t \to 0 \), the update process converges to a differential equation:
	
	\begin{equation}
		\frac{d\mathbf{W}^{(l)}}{dt} + \nabla_{\mathbf{w}} H^{(l)} = 0.
	\end{equation}
These differential equations can be expanded to their explicit form for each layer, where we see the weight update is formed from the outer product of the input and the entropy gradient with respect to knowledge:
	
	\begin{equation}
		\frac{d\mathbf{W}^{(1)}}{dt} = -\nabla_{\mathbf{z}} H^{(1)} \otimes  \mathbf{X} 
	\end{equation}
	
	\begin{equation}
		\frac{d\mathbf{W}^{(l)}}{dt} = -\nabla_{\mathbf{z}} H^{(l)} \otimes  \mathbf{D}^{(l-1)}   \quad \text{for } l > 1
	\end{equation}
	
For the first layer, this input is the raw data  $\mathbf{X}$, while for subsequent layers, the input becomes the decision probabilities $\mathbf{D}^{(l-1)}$ from the previous layer. This means that weight changes arise locally from how inputs contribute to reducing entropy through the entropy gradient. As a result, each layer autonomously adjusts its weights to optimize information flow based solely on its own input and knowledge state, without requiring any backpropagation from deeper layers.
	
	This reveals a key property of SKA: the learning trajectory is not merely a sequence of adjustments but an evolution governed by differential equations, akin to natural physical processes such as diffusion or energy dissipation. Consequently, \( \eta \) is no longer an arbitrary choice but a fundamental timescale determining the smoothness and speed of structured knowledge accumulation.

	\subsubsection{Evidence for Time-Invariant Behavior}
	
	To validate this reinterpretation, we conducted a series of experiments with different values of $\eta$ (time step) and corresponding adjustments to $K$ (number of steps) such that their product $\eta \times K$ remained constant at 0.5. Remarkably, we observed that the SKA system follows the same learning trajectory regardless of the specific values of $\eta$ and $K$, as long as their product remains fixed.
	
\begin{figure}[htp!]
		\centering
		\includegraphics[width=1.0\textwidth]{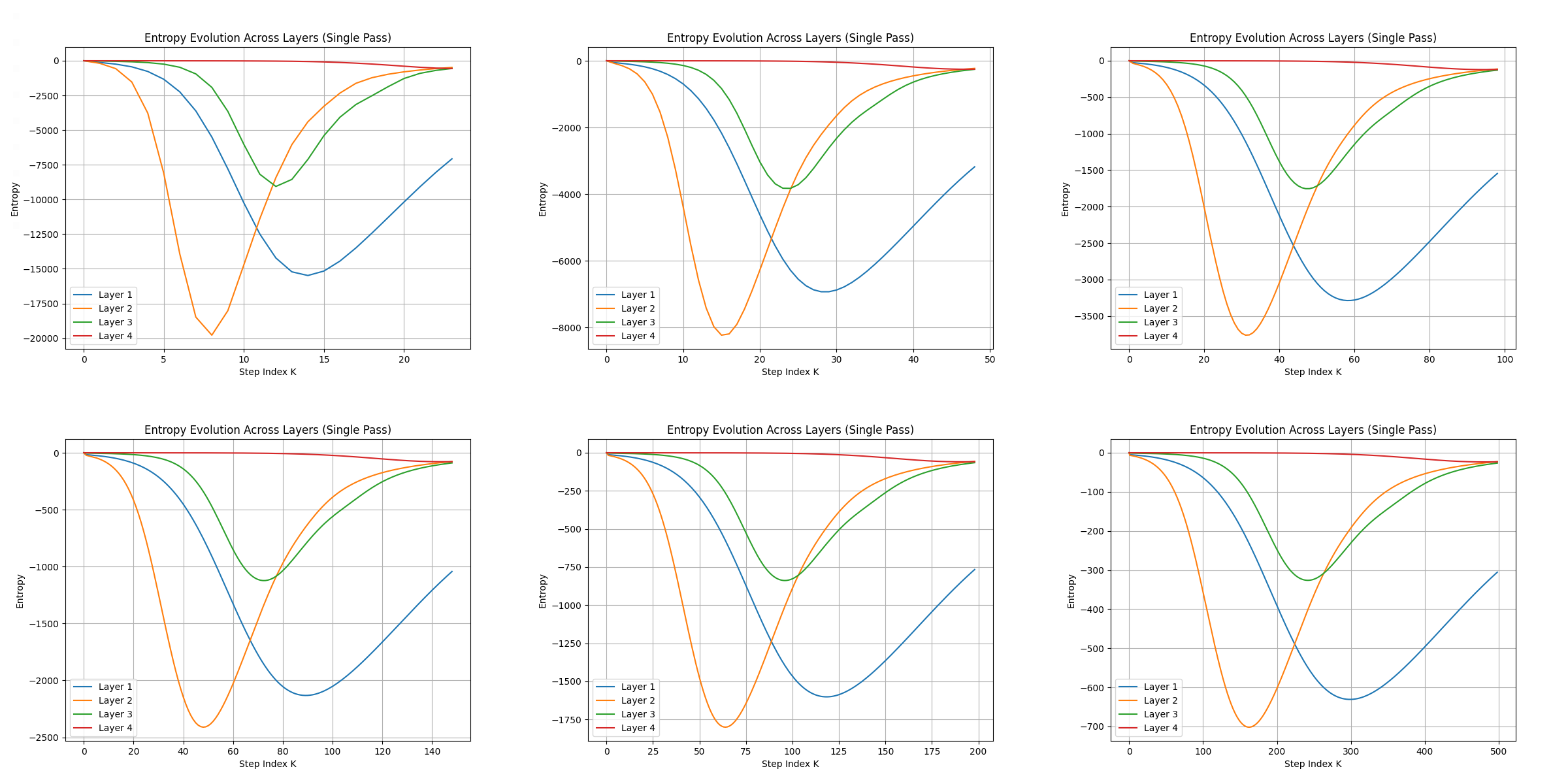}
		\caption{Comparison of entropy evolution patterns across different time step configurations. Despite different discretization granularity, the overall trajectories remain identical when $\eta \times K = 0.5$, demonstrating the time-invariant nature of SKA learning.}
		\label{fig:entropy_history_time_invariant}
	\end{figure}

\begin{figure}[htp!]
		\centering
		\includegraphics[width=1.0\textwidth]{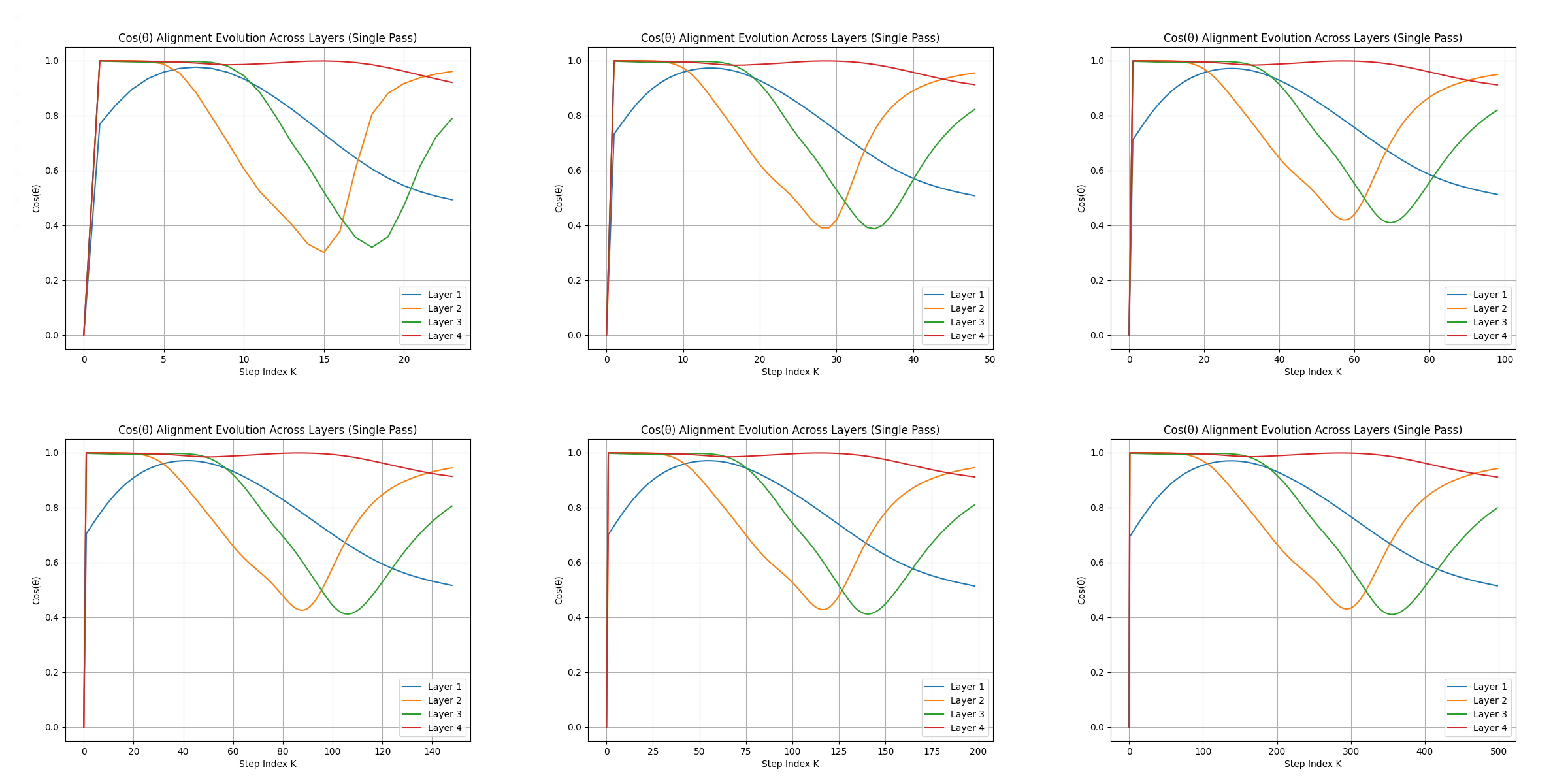}
		\caption{Comparison of cosine evolution patterns across different time step configurations. Despite different discretization granularity, the overall trajectories remain identical when $\eta \times K = 0.5$, demonstrating the time-invariant nature of SKA learning.}
		\label{fig:cosine_history_time_invariant}
	\end{figure}

	Figure~\ref{fig:entropy_history_time_invariant} and Figure~\ref{fig:cosine_history_time_invariant} compares the entropy evolution and cosine alignment patterns across different configurations:
	\begin{itemize}
		\item $\eta = 0.02$, $K = 25$
		\item $\eta = 0.01$, $K = 50$
		\item $\eta = 0.005$, $K = 100$
		\item $\eta = 0.0033$, $K = 150$
		\item $\eta = 0.0025$, $K = 200$
		\item $\eta = 0.001$, $K = 500$
	\end{itemize}
	
	The striking similarity in these patterns, despite the 20-fold difference in discretization granularity, provides compelling evidence that the SKA learning process is truly time-invariant when viewed through the lens of total integration time $T = \eta \times K = 0.5$.
	
\FloatBarrier	
\FloatBarrier	
This time-invariant behavior is characteristic of well-behaved differential equations and strongly suggests that SKA is modeling a genuine continuous-time process rather than a discrete optimization procedure.
	
\subsection{The Characteristic Time}
	
\subsubsection{Definition and Empirical Validation}
Our experiments reveal the existence of a characteristic time $T = \eta \times K = 0.5$ for the neural network architecture used in our studies. This characteristic time represents the intrinsic timescale over which the SKA system completes its natural learning dynamics.
The value $T = 0.5$ appears to be a fundamental property of the specific network configuration (4 layers with sizes [256, 128, 64, 10] processing MNIST data), as it consistently emerges across different discretization schemes. This suggests that each neural architecture may have its own characteristic timescale that governs its learning dynamics.
	
	\subsubsection{Physical Interpretation}
	
	The characteristic time can be understood as the time required for knowledge to properly structure itself across the network's layers. It is analogous to natural timescales in physical systems, such as:
	
	\begin{itemize}
		\item The period of oscillation in mechanical systems
		\item The time constant in RC circuits
		\item The residence time for a gas in a container with a constant flow
	\end{itemize}
	
	In a neural system context, the characteristic time represents how long it takes for knowledge to accumulate, organize, and reach equilibrium across the network hierarchy. This time depends on the network's capacity to store and structure information relative to the complexity of the patterns being learned.
	
	Using the fluid dynamics analogy, the characteristic time can be viewed as a "residence time" for information flow through the network:
	
	\begin{equation}
		T = \frac{\text{knowledge capacity}}{\text{knowledge flow}}
	\end{equation}
	
	This perspective aligns with biological neural systems, which also operate on intrinsic timescales determined by their structural properties.
	
	\subsection{The Knowledge Flow}
	
	\subsubsection{Definition}

	The concept of knowledge flow  emerges naturally from our reinterpretation of the learning rate \( \eta \) as a time step \( \Delta t \). Defined analogously to the concept of flow in fluid dynamics, the knowledge flow rate quantifies how rapidly knowledge is accumulated or transmitted across the layers of the neural network over time.
	
	Formally, we define the knowledge flow \( \Phi(t) \) at any step \( t \) as the instantaneous rate at which knowledge changes with respect to time:
	
	\begin{equation}
		\Phi(t) = \frac{dZ}{dt}
	\end{equation}
	
	where \( Z(t) \) represents the cumulative knowledge at step \( t \). In the discrete SKA framework, this becomes:
	
	\begin{equation}
		\Phi_k = \frac{Z_{k} - Z_{k-1}}{\Delta t}.
	\end{equation}

	\subsubsection{The Evolution of Knowledge Flow and Entropy Minimization}
	
\begin{figure}[htp!]
		\centering
		\includegraphics[width=1.0\textwidth]{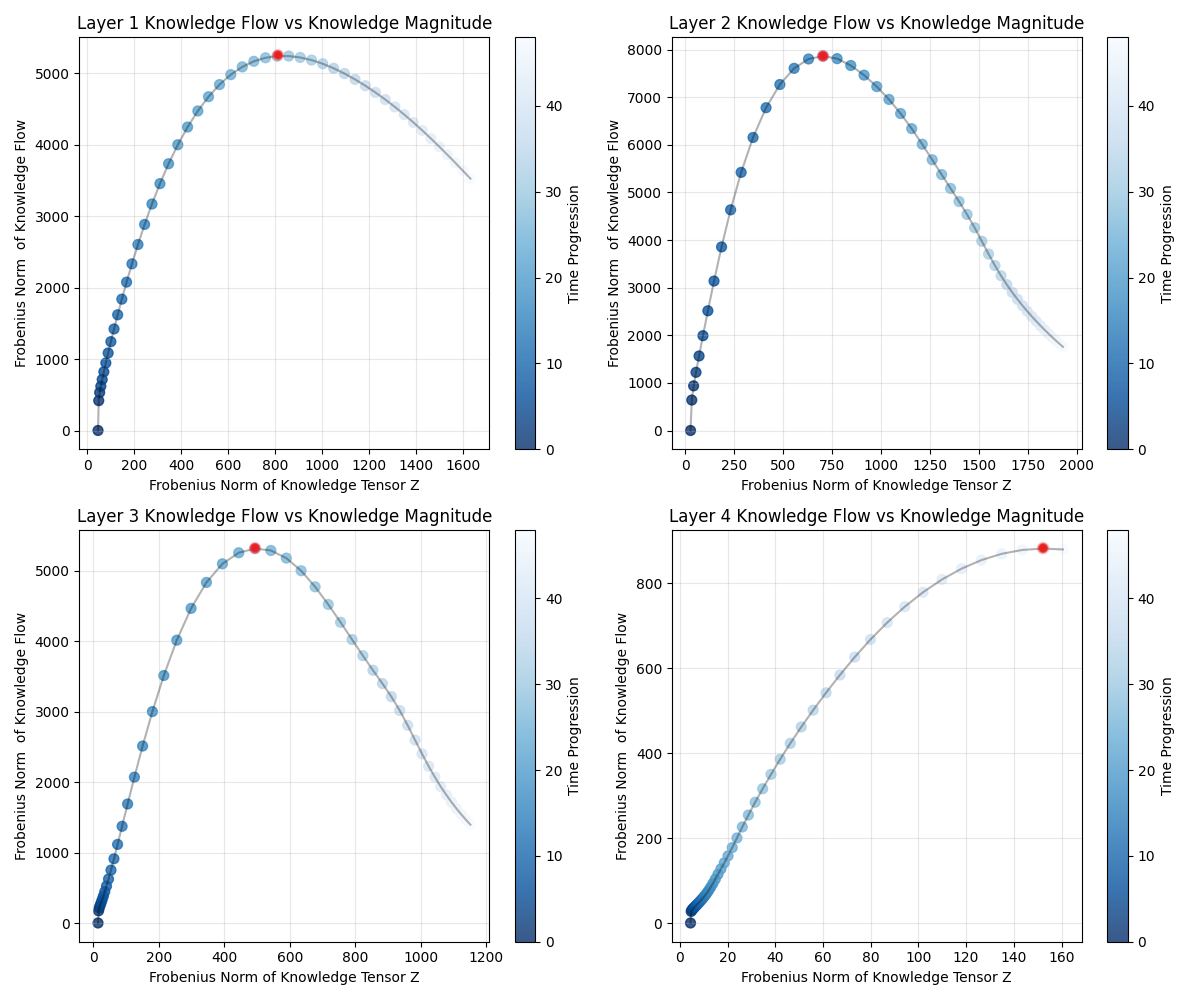}
		\caption{Knowledge Flow versus the Frobenius norm of the knowledge tensor for each layer. The color gradient represents training progression (darker blue points occur earlier in training). Red dots mark the entropy minima for each layer. The experiment is conducted within the valid characteristic time scale ($T = 0.5$).}
		\label{fig:tensor_flow_vs_knowledge}
	\end{figure}

\begin{figure}[htp!]
		\centering
		\includegraphics[width=1.0\textwidth]{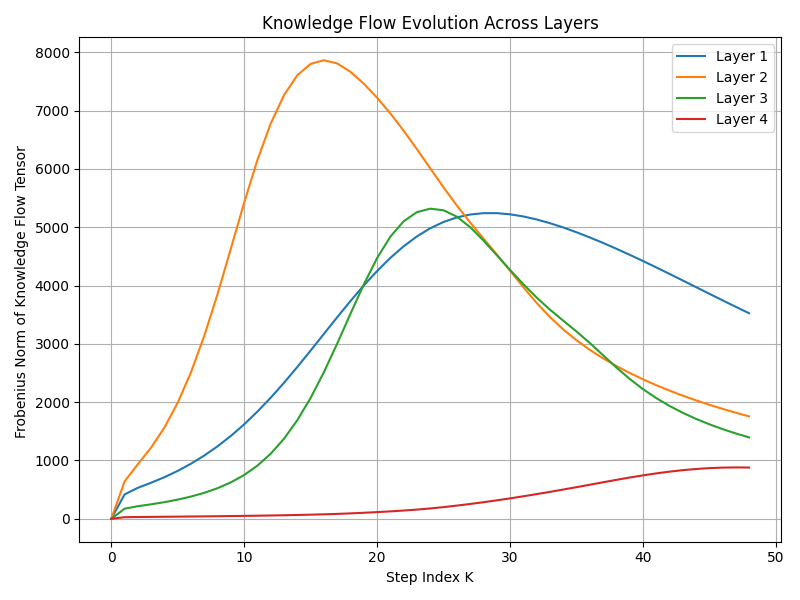}
		\caption{Knowledge Flow evolution over discrete steps K. The experiment is conducted within the valid characteristic time scale ($T = 0.5$), ensuring that knowledge flow convergence accurately represents structured knowledge accumulation completion.}
		\label{fig:knowledge_flow_evolution}
	\end{figure}

	A key observation in our study is the alignment between knowledge flow and entropy minimization across different layers. Figures~\ref{fig:tensor_flow_vs_knowledge} and~\ref{fig:knowledge_flow_evolution} visualize this relationship, respectively showing:
	
	1. The Frobenius norm of knowledge flow versus knowledge tensor  magnitude for each layer.
	
	2. The temporal evolution of knowledge flow across layers.
	
	Our experimental results strictly adhere to the characteristic time scale ($T = 0.5$), ensuring that observed convergence phenomena are aligned with structured knowledge accumulation completion.
	
	\subsubsubsection{Figure 4: Knowledge Flow vs Knowledge Tensor}
	Figure~\ref{fig:tensor_flow_vs_knowledge} plots knowledge flow against the Frobenius norm of the knowledge tensor. The entropy minima (red dots) mark the points where structured knowledge accumulation reaches an optimal state. As expected, Layers 1–3 exhibit clear peaks in knowledge flow that align with entropy minimization. In contrast, Layer 4 follows a gradual increase without a distinct peak, indicating a transition from feature extraction to decision-making.
	
\FloatBarrier	
	\subsubsubsection{Figure 5: Temporal Evolution of Knowledge Flow}
	In Figure~\ref{fig:knowledge_flow_evolution}, we examine the temporal evolution of knowledge flow across layers. The experiment is strictly within the correct time scale ($T = 0.5$), ensuring that knowledge flow convergence is meaningful and accurately reflects structured knowledge accumulation completion.
	
	Key observations include:
	
	- Layer 2 experiences the most rapid increase, peaking at $ K \approx 16 $.
	
	- Layer 3 and Layer 1 reach their peak knowledge flow later, around $ K \approx 24 $ and $ K \approx 28 $, respectively.
	
	- Layer 4 exhibits a gradual, lower trajectory, reinforcing its role in classification rather than feature extraction.
	
	- Convergence of knowledge flow across layers marks the structured knowledge accumulation completion, in accordance with entropy convergence (Figure~\ref{fig:entropy_history_time_invariant}).
	
\FloatBarrier	
	\subsection{The Tensor Net Function}
	
	\subsubsection{Definition and Significance}
	
	A critical insight in the SKA framework is the \textbf{Tensor Net function}, which measures the balance between decision probability and entropy gradient with respect to knowledge. For each layer $l$ at step $K$, the Tensor Net function is derived directly from the governing differential equations of the SKA framework:
	
	\begin{equation}
		\text{Net}^{(l)}_K = \sum_k  (\mathbf{D}^{(l)}_k - \nabla_z \mathbf{H}^{(l)}_k) \cdot \Delta \mathbf{Z}^{(l)}_k 
	\end{equation}
	
	where $\mathbf{D}^{(l)}_k$ represents the decision probability tensor, $\nabla_z \mathbf{H}^{(l)}_k$ is the gradient of entropy with respect to knowledge, and $\Delta \mathbf{Z}^{(l)}_k$ is the change in knowledge tensor between steps. This function captures the inner product between knowledge changes and the difference between decision probabilities and entropy gradient.  

This approach comes intuitively from the differential equation $\nabla_z \mathbf{H}^{(l)} + \frac{1}{\ln 2}\mathbf{z}^{(l)} \odot \mathbf{D'}^{(l)} = 0$ that describes the SKA framework in the time-continuous model. Given the entropy gradient $\nabla_z \mathbf{H}^{(l)}_k = \frac{-1}{\ln 2}\mathbf{z}^{(l)}_k \odot \mathbf{D'}^{(l)}_k$, the Tensor Net function gives a well defined measure of the alignment between the knowledge evolution and the decision probabilities combined with the entropy gradient.

The condition \(\mathbf{D}^{(l)}_k = \nabla_z \mathbf{H}^{(l)}_k\) describes the extrema of the Tensor Net function, which indicates the balance point where decision probabilities and entropy gradients intersect. The zero-crossing points (\( \text{Net}^{(l)}_K = 0\)) denote the boundary separating unstructured from structured learning and serves as a principled criterion for determining when knowledge accumulation starts. Mathematically, this condition is given as $\int \mathbf{D}^{(l)} \, dz = \mathbf{H}^{(l)}$.

\subsubsection{Empirical Observations}
\begin{figure}[htp!]
		\centering
		\includegraphics[width=1.0\textwidth]{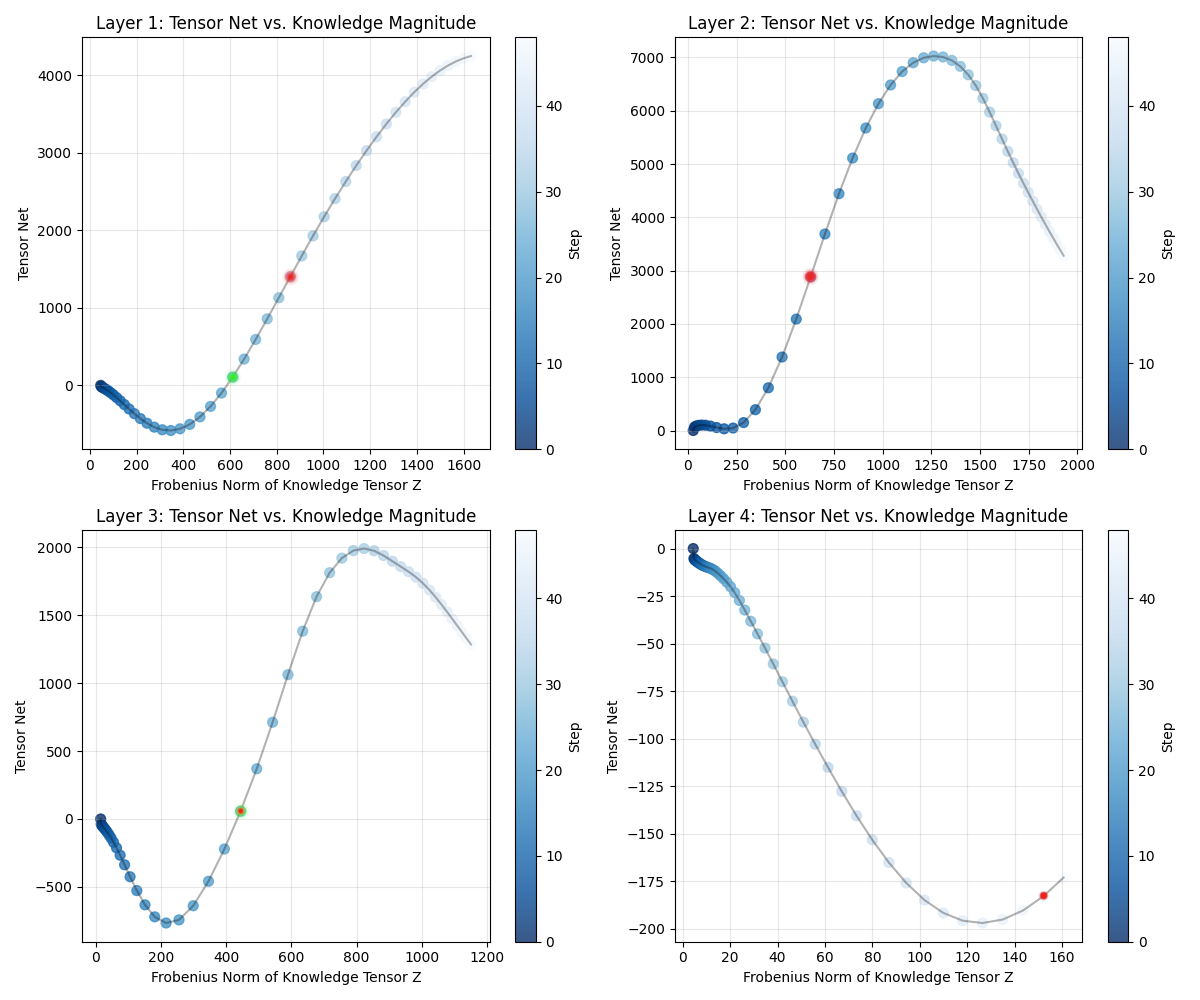}
		\caption{The red marks illustrate the entropy minima present for each layer. The green dots represent the zero-crossing points where $\int \mathbf{D}^{(l)} \, dz = \mathbf{H}^{(l)}$. Each layer shows different behaviors that correspond to its position in the hierarchy. The earlier layers have smooth transitions, whereas, the output layer (Layer 4) has a negative trajectory which is attributed to its classification task.}
		\label{fig:tensor_net_vs_knowledge}
	\end{figure}

\paragraph{Tensor Net vs Knowledge Tensor:}
The connection between the Tensor Net function and the knowledge magnitude, given specifically by the Frobenius norm of the knowledge tensor \(\mathbf{Z}^{(l)}\), provides important information about how each layer learns. This relationship for the four layers of the network is shown in Figure~\ref{fig:tensor_net_vs_knowledge}.

	Several notable patterns emerge from this visualization:
	
	\begin{itemize}
		\item \textbf{Layer-Specific Trajectories}: Every layer demonstrates its own trajectory, with Layer 1 steadily growing after an initial drop, Layer 2 exhibiting a classic parabola with a peak around 1250 units, Layer 3 displaying a similar curve with an earlier peak, and Layer 4 consistently showing negative results.

		\item \textbf{Entropy Minima Alignment}: The red dots marking entropy minima show varying relationships with the Tensor Net function across layers. In Layer 1, the entropy minimum occurs after the zero-crossing point, indicating that entropy optimization continues even as the Tensor Net becomes increasingly positive. In Layer 2, it occurs early in the positive trajectory, while in Layer 4, it appears at the highest knowledge magnitude where the curve begins to turn upward.
		
		\item \textbf{Zero-Crossing Points}: The green dots indicate where the integral of decision probabilities over knowledge equals the integral of entropy gradient over knowledge ($\int \mathbf{D}^{(l)} \, dz = \mathbf{H}^{(l)}$). These zero-crossing points represent fundamental equilibrium states in the information flow. Layer 1's zero-crossing occurs at approximately 600 units, Layer 3's at about 450 units, establishing boundaries between different regimes of knowledge organization. It is at this point that structured knowledge accumulation begins.
		
		\item \textbf{Distinctive Pattern in Layer 4}: Unlike the other layers, Layer 4 (the output layer) shows uniquely negative Tensor Net values throughout most of its trajectory, with values decreasing as knowledge magnitude increases. This distinctive behavior reflects its specialized role in converting internal representations to classification decisions.
	\end{itemize}

    This analysis shows that each one of the layers possesses a distinct knowledge magnitude profile in relation to the evolution of the Tensor Net function. The existence of entropy minima and zero-crossing identification for each layer at
different magnitudes of knowledge suggests a hierarchical, layer-wise knowledge systematization process. 

In the context of the SKA framework, the Tensor Net function, which is derived from the governing differential equations, does two jobs simultaneously. It provides us with an explanation of structured knowledge dynamics and serves as
a practical tool for tracking learning progress. It is crucial to note that the zero-crossing of the Tensor Net function marks a vital boundary, which is the point when the system changes from mere disorganized alteration to meaningful structuring of knowledge. On the other hand, the convergence of combined entropy with knowledge flow indicates the system equilibrium state, where the accumulation of structured knowledge is complete. In this condition, the network has fully structured its internal representations, achieving a balanced and efficient distribution of information across its hierarchical architecture.
	
\FloatBarrier	

\begin{figure}[htp!]
	\centering
	\includegraphics[width=1.0\textwidth]{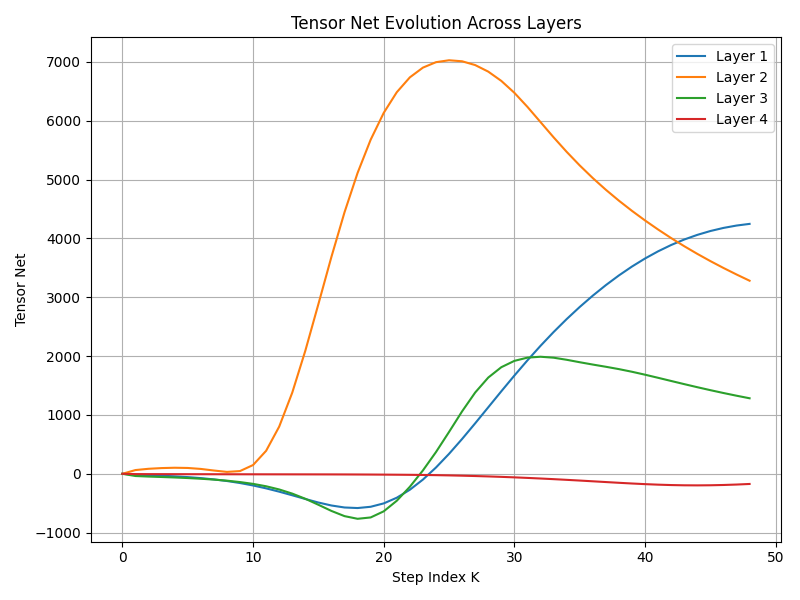}
	\caption{Tensor Net evolution over discrete steps $K$ within the characteristic time scale ($T = 0.5$). Zero-crossing points indicate transitions from unstructured to structured learning across different layers of the network.}
	\label{fig:tensor_net_evolution}
\end{figure}

\paragraph{Tensor Net Temporal Evolution:}
	The temporal evolution of the Tensor Net function reveals distinct layer-specific behaviors across forward steps, as shown in Figure~\ref{fig:tensor_net_evolution}. Each layer exhibits a unique trajectory, with particularly notable zero-crossing points marking the transition from unstructured to structured learning. Layers 1 and 3 cross the zero threshold at different steps (approximately $K=24$ for Layer 1 and $K=23$ for Layer 3), indicating layer-specific timing for the onset of meaningful knowledge accumulation. Layer 2 begins with positive values and shows the most dramatic peak, while Layer 4 maintains consistently low values throughout the process. These diverse patterns reflect the hierarchical nature of knowledge organization, with earlier layers showing more pronounced changes in the balance between decision probabilities and entropy gradients, while the output layer remains in a specialized state focused on classification.

\FloatBarrier

\subsection{Variational Analysis of the SKA Framework}

In our previous work \cite{mahi2025ska}, we observed that "layer-wise entropy converges to an equilibrium state where knowledge accumulation stabilizes across hierarchical representations" and suggested "the possible existence of a fundamental law governing SKA-based neural networks." The time-invariant properties and characteristic timescales demonstrated in the current paper further support this hypothesis.

We can now formalize this intuition through the lens of variational principles, providing a mathematical foundation for the self-organized learning behavior observed in SKA. This analysis reveals that the equilibrium states and structured knowledge accumulation observed empirically emerge naturally from a more fundamental variational principle.

Here, we demonstrate how entropy H relates to a Lagrangian formulation and show that SKA dynamics naturally satisfy the Euler–Lagrange equation.

\subsubsection{Entropy and Lagrangian Formulation}

In the SKA framework, we have defined entropy as:
\begin{equation}
	H = -\frac{1}{\ln 2} \int z \, dD
\end{equation}

This can be reformulated in terms of a time integral by recognizing that $dD = \dot{D} \, dt$, giving us:
\begin{equation}
	H = -\frac{1}{\ln 2} \int z \, \dot{D} \, dt.
\end{equation}

This immediately suggests a Lagrangian density of the form:
\begin{equation}
	\mathcal{L}(z, \dot{D}, t) = -z \cdot \dot{D}.
\end{equation}

Thus, the entropy of the system can be expressed as an action integral:
\begin{equation}
	H = \frac{1}{\ln 2} \int \mathcal{L} \, dt.
\end{equation}

Since $D = \sigma(z) = \frac{1}{1 + e^{-z}}$, the derivative of $D$ with respect to time can be expressed as:

\begin{equation}
	\dot{D} = \frac{dD}{dz} \cdot \dot{z} = \sigma(z)(1 - \sigma(z)) \cdot \dot{z}
\end{equation}

This transforms our Lagrangian into:

\begin{equation}
	\mathcal{L}(z, \dot{z}, t) = -z \cdot \sigma(z)(1 - \sigma(z)) \cdot \dot{z}
\end{equation}

For simplicity, let $S = \sigma(z)$, giving us:

\begin{equation}
	\mathcal{L}(z, \dot{z}, t) = -z \cdot S(1 - S) \cdot \dot{z}.
\end{equation}

\subsubsection{Computing the Euler–Lagrange Equation}

The Euler–Lagrange equation provides the condition for extremizing the action integral:

\begin{equation}
	\frac{d}{dt}\left(\frac{\partial \mathcal{L}}{\partial \dot{z}}\right) - \frac{\partial \mathcal{L}}{\partial z} = 0
\end{equation}

Computing the partial derivatives:

\begin{equation}
	\frac{\partial \mathcal{L}}{\partial \dot{z}} = -z \cdot S(1 - S)
\end{equation}

\begin{equation}
	\frac{\partial \mathcal{L}}{\partial z} = -\dot{z} \cdot \left[S(1 - S) + z \cdot \frac{d}{dz}(S(1 - S))\right]
\end{equation}

The derivative $\frac{d}{dz}(S(1 - S))$ can be calculated as:

\begin{equation}
	\frac{d}{dz}(S(1 - S)) = S'(1 - S) - S \cdot S' = S(1 - S)(1 - 2S)
\end{equation}

Therefore:

\begin{equation}
	\frac{\partial \mathcal{L}}{\partial z} = -\dot{z} \cdot \left[S(1 - S) + z \cdot S(1 - S)(1 - 2S)\right]
\end{equation}

Now we compute $\frac{d}{dt}\left(\frac{\partial \mathcal{L}}{\partial \dot{z}}\right)$:

\begin{equation}
	\frac{d}{dt}\left(-z \cdot S(1 - S)\right) = -\left[\dot{z} \cdot S(1 - S) + z \cdot \frac{d}{dt}(S(1 - S))\right]
\end{equation}

Since $\frac{d}{dt}(S(1 - S)) = \frac{d}{dz}(S(1 - S)) \cdot \dot{z} = S(1 - S)(1 - 2S) \cdot \dot{z}$, we have:

\begin{equation}
	\frac{d}{dt}\left(-z \cdot S(1 - S)\right) = -\left[\dot{z} \cdot S(1 - S) + z \cdot S(1 - S)(1 - 2S) \cdot \dot{z}\right]
\end{equation}

Substituting these expressions into the Euler-Lagrange equation:

\begin{align}
	&-\left[\dot{z} \cdot S(1 - S) + z \cdot S(1 - S)(1 - 2S) \cdot \dot{z}\right] + \dot{z} \cdot \left[S(1 - S) + z \cdot S(1 - S)(1 - 2S)\right]\\
	&= -\dot{z} \cdot S(1 - S) - z \cdot S(1 - S)(1 - 2S) \cdot \dot{z} + \dot{z} \cdot S(1 - S) + z \cdot S(1 - S)(1 - 2S) \cdot \dot{z}\\
	&= 0
\end{align}

\paragraph{Interpretation:}
This remarkable result—where the Euler–Lagrange equation reduces to $0=0$—has profound implications for the SKA framework:

\begin{enumerate}
	\item \textbf{Natural Optimality}: The SKA dynamics naturally satisfy the principle of least action without requiring additional constraints. This indicates that the learning trajectory in SKA is inherently optimal from a
variational standpoint.
	
	\item \textbf{Self-Consistency}: The fact that the equation is identically satisfied demonstrates the mathematical consistency of the SKA framework. The system evolves along paths that naturally minimize the action integral.
	
	\item \textbf{Connection to Physical Systems}: This result strengthens the parallel between SKA and physical systems governed by variational principles, further supporting our interpretation of learning as a continuous-time process following natural laws.
\end{enumerate}

This variational analysis provides strong theoretical support for the time-invariant properties observed in SKA and reinforces our reframing of neural learning as a continuous process governed by principles analogous to those in physical systems. Rather than emerging from discrete optimization steps, the learning trajectory unfolds as a smooth evolution that naturally satisfies variational principles—just as physical systems follow paths of least action.

A key property of the SKA Lagrangian is its explicit time-independence: \( \frac{\partial \mathcal{L}}{\partial t} = 0 \). This implies that the learning dynamics are driven entirely by internal variables, not by time itself. This time-independence of the Lagrangian explains the scaling behavior observed in our entropy experiments (Figure~\ref{fig:entropy_history_time_invariant}), where entropy values scale proportionally with the time step $\eta$ while preserving identical trajectory shapes. This mathematical relationship confirms that $T = \eta \times K = 0.5$ represents a fundamental characteristic time of the system, independent of discretizations.

Since all trajectories satisfy the Euler--Lagrange equation (as shown by the identity \( 0 = 0 \)), the actual learning path taken by the network is fully determined by boundary conditions—such as initialization, architecture, and data distribution. This explains why distinct initializations can lead to different, yet equally structured and effective, knowledge representations. The empirical timescale \( \eta \times K = 0.5 \) may thus reflect an emergent boundary condition shaped by the system’s internal configuration.

The formulation presented here establishes a rigorous variational foundation for structured knowledge accumulation, defining SKA learning as a trajectory that systematically minimizes the entropy action across the network’s layers. This principle unifies our empirical observations of characteristic timescales, entropy flow convergence, and the Tensor Net function’s role in marking phase transitions.

\paragraph{Principle of Entropic Least Action:}
The optimal learning trajectory in the SKA framework is precisely the one that minimizes the entropy action integral, expressed as:
	\[
	H = \frac{1}{\ln 2} \int \mathcal{L}(z, \dot{z}, t) \, dt,
	\quad \text{with} \quad
	\mathcal{L}(z, \dot{z}, t) = -z \cdot \sigma(z)(1 - \sigma(z)) \cdot \dot{z}
	\]
While this principle governs all possible learning trajectories, the particular path through knowledge space is uniquely determined by boundary conditions.
	
\subsubsection{Tensor Net and Lagrangian Formulation}

The Tensor Net function, initially defined as a measure of the balance between decision probabilities and entropy gradients, reveals a deeper connection to the variational principles underlying SKA. When we express the Tensor Net function in its continuous form:

\begin{equation}
	\text{Net} = \int (D - \nabla_z H) \, dz
\end{equation}

We can relate this directly to our Lagrangian formulation. Given that $D = \sigma(z)$ and $\nabla_z H = -\frac{1}{\ln 2}z \cdot \sigma(z)(1-\sigma(z))$, the Tensor Net function can be rewritten as:

\begin{equation}
	\text{Net} = \int \left(\sigma(z) + \frac{1}{\ln 2}z \cdot \sigma(z)(1-\sigma(z))\right) \cdot \dot{z} \, dt
\end{equation}

Recognizing that our Lagrangian is defined as $\mathcal{L} = -z \cdot \sigma(z)(1-\sigma(z)) \cdot \dot{z}$, we can express the relationship as:

\begin{equation}
	\text{Net} = \int \left(\sigma(z) \cdot \dot{z} - \frac{1}{\ln 2}\mathcal{L}\right) \, dt
\end{equation}

This relationship reveals that the Tensor Net function integrates the difference between the product of the decision probability and the knowledge flow, and the Lagrangian. When we compute the Euler-Lagrange equation for this integrated expression, we again arrive at the identity $0 = 0$, just as we did with our original Lagrangian. This remarkable result shows that the Tensor Net function satisfies the same fundamental variational principle as the Lagrangian formulation.

The zero-crossing points of the Tensor Net function, which we defined as marking the transition to structured knowledge accumulation, now gain deeper theoretical justification. When $\text{Net} = 0$, we have:
\begin{equation}
	\int \sigma(z) \cdot \dot{z} \, dt = \frac{1}{\ln 2} \int \mathcal{L} \, dt =  H
\end{equation}

The fact that both approaches lead to the identity $0 = 0$ suggests we have identified two complementary aspects of the same underlying principle governing neural learning dynamics.

	\subsection{Implications and Applications}
	
	\subsubsection{Theoretical Implications}
	The finding of characteristic time property, Tensor Net function, and knowledge flow convergence has large scale theoretical impact due to its:

    \begin{itemize}
		
		\item \textbf{Neural Learning as a Physical Process}: SKA reveals that neural learning is a physical phenomenon, governed by natural laws and intrinsic timescales—not arbitrary optimization schemes. This physical interpretation is further strengthened by the variational principles underlying the system's dynamics.
		
		\item \textbf{Bridge to Dynamical Systems Theory}: The time-invariance properties of SKA have an immediate relation to dynamical systems theory, which benefits neural learning from a structural-analysis viewpoint.

        \item \textbf{Emergence of Stopping Criteria}: For the Tensor Net function, the zero-crossing provides an emergent criterion for determining when structured learning starts while entropy and knowledge flow convergence signal its completion.
		
		\item \textbf{Fundamental Coupling of Knowledge and Entropy}: The association of knowledge flow convergence with entropy equilibrium validates the coupling of knowledge accumulation and entropy reduction and indicates a primary attribute of information-organizing systems. This coupling emerges naturally from the principle of entropic least action.

        \item \textbf{Self-Solving System}: One of the properties of SKA is that it serves as the numerical solver for the difference equations that describe it. Knowledge accumulation leads to the natural solution of entropy gradient equations by the forward dynamics of the network which create a self-referential system where implementation and solution method become one and the same.
	\end{itemize}

	\subsubsection{Practical Applications}
These information also provide practical benefits for neural network implementation:
	
    \begin{itemize}

    \item \textbf{Adaptive Time Stepping}: Use of adaptive time stepping methods from numerical integration may speed up SKA learning.

		\item \textbf{Efficient Training Schedules}: Knowledge of the characteristic times enables optimum selection of learning parameters relative to the network design.  

		\item \textbf{Architectural Optimization}: Networks can be built with certain characteristic times for certain organs to achieve various applications.

        \item \textbf{Resource-Efficient Implementation}: The predictable timescales of SKA, combined with its forward-only nature, allows more efficient hardware implementations which eliminate the computational overhead of backpropagation.

		\item \textbf{Flow-Based Monitoring}: Monitoring knowledge flow during training gives natural metrics for convergence and system stability, which could allow for early stopping without arbitrary thresholds. These metrics reflect the system's progress toward entropic action minimization.

	\end{itemize}

\section{Conclusion and Future Research Directions}
This work has significantly extended the SKA framework with the following three contributions: the introduction of the Tensor Net function for quantifying the occurrence of structured learning, the characteristic time property as a natural quality of neural learning dynamics, and the convergence of knowledge flow as a reliable signal of systemic optimization. 

A major finding of this research is the SKA formulation as a continuous-time learning model grounded in first principles. This formulation is represented by the following differential equations, which describe the knowledge distribution entropy, its gradient behavior, and the temporal evolution of neural weights.

\begin{equation}
	H = -\frac{1}{\ln 2} \int z \, dD,
\end{equation}

\begin{equation}
	\nabla_{\mathbf{z}} H^{(l)} + \frac{1}{\ln 2} \mathbf{z}^{(l)} \odot \mathbf{D}^{\prime(l)} = 0,
\end{equation}

\begin{equation}
	\frac{d\mathbf{W}^{(l)}}{dt} + \nabla_{\mathbf{w}} H^{(l)} = 0.
\end{equation}

These equations provide the mathematical framework of learning dynamics in continuous time, moving beyond discrete optimization and toward a formulation governed by entropy flow and natural timescales.

Our obtained results reshape the established perspective of learning in neural networks by showcasing the time-invariant properties and physically meaningful behavior of knowledge evolution.  This work also bridges the domains of computational learning and natural processes by positioning SKA within the framework of dynamical systems, and offers a new lens through which AI can be grasped and implemented. The principle of entropic least action established in this work provides a fundamental variational foundation that unifies our empirical observations and explains why neural learning follows paths that minimize entropic action, analogous to how physical systems minimize action integrals.

One of the most significant properties of SKA is its self-solving nature—wherein the network itself acts as the numerical solver for its governing equations. This self-referential mechanism reflects how physical systems naturally evolve according to their own internal laws, positioning SKA as a more principled and theoretically grounded alternative to conventional gradient-based learning.

\subsection{Future Research Directions}

Building on the contributions presented in this work, future research will aim to deepen and expand the SKA framework through the following directions:

\begin{itemize}
    \item Derivation of characteristic time scales across diverse network architectures and learning environments, and their relationship to the principle of entropic least action.
    \item Development of theoretical models to predict and analyze Tensor Net function behavior during different learning phases.
    \item Integration of adaptive time-stepping techniques to optimize learning efficiency and convergence speed.
    \item Extension of SKA to support complex architectures, including recurrent networks, graph neural networks, and large-scale transformer models.
    \item Mathematical exploration of the relationship between knowledge flow convergence and the emergence of optimal information structures through variational analysis.
    \item Design of hardware architectures specifically tailored to leverage SKA’s forward-only and time-invariant properties for efficient learning.
\end{itemize}

Through these avenues, we aim to position SKA not merely as an alternative learning strategy, but as a foundational framework for modeling, understanding, and realizing learning processes in both artificial and biological systems.

%%%%%%%%%%%%%%%%%%%%%%%%%%%%%%%%%%%%%%%%%%%%%%%%%%%%%%%%%%%%%%%%%%
\section*{Data Availability}
The data that support the findings of this study are available from the corresponding author upon reasonable request.
%%%%%%%%%%%%%%%%%%%%%%%%%%%%%%%%%%%%%
\section*{Competing Interests}
The authors declare that there are no competing interests.

\section*{Author Contributions}
The author confirms sole responsibility for all aspects of this work, including conceptualization, methodology, analysis, writing, and manuscript preparation.

\section*{Ethics Declaration}
Not applicable.

\section*{Funding Declaration}
This research received no specific grant from any funding agency in the public, commercial, or not-for-profit sectors.

%%%%%%%%%%%%%%%%%%%%%%%%%%%%%%%%%%%%%%%%%%%%%%%%%%%%%%%%%%%%%%%%%%%%%%%%%%%%%%%%%%%%%%%%%%%%%%%%%%%%%%%
\bibliographystyle{unsrt}
%\bibliography{references}  %%% Uncomment this line and comment out the ``thebibliography'' section below to use the external .bib file (using bibtex) .


\begin{thebibliography}{1}

\bibitem{mahi2025ska}
Mahi, B., 2025. \textit{Structured Knowledge Accumulation: An Autonomous Framework for Layer-Wise Entropy Reduction in Neural Learning}. arXiv preprint.

\bibitem{feulner2025}
Feulner, B., Perich, M.G., Miller, L.E., Clopath, C. and Gallego, J.A., 2025. \textit{A neural implementation model of feedback-based motor learning}. \textit{Nature Communications}, 16(1), p.1805.

\bibitem{terres2024}
Terres-Escudero, E.B., Del Ser, J., Martínez-Seras, A. and Garcia-Bringas, P., 2024. \textit{On the Robustness of Fully-Spiking Neural Networks in Open-World Scenarios using Forward-Only Learning Algorithms}. arXiv preprint arXiv:2407.14097.

\bibitem{chen2018neural}
Chen, R.T.Q., Rubanova, Y., Bettencourt, J. and Duvenaud, D., 2018. \textit{Neural ordinary differential equations}. In: Advances in Neural Information Processing Systems (NeurIPS), pp.6571--6583.

\bibitem{lillicrap2016random}
Lillicrap, T.P., Cownden, D., Tweed, D.B. and Akerman, C.J., 2016. \textit{Random synaptic feedback weights support error backpropagation for deep learning}. Nature Communications, 7, p.13276.

\bibitem{balduzzi2017mechanistic}
Balduzzi, D., 2017. \textit{The mechanistic and thermodynamic structure of learning algorithms}. arXiv preprint arXiv:1705.08020.

\bibitem{rumelhart1986learning}
Rumelhart, D.E., Hinton, G.E. and Williams, R.J., 1986. \textit{Learning representations by back-propagating errors}. Nature, 323, pp.533--536.

\bibitem{whittington2017approximation}
Whittington, J.C.R. and Bogacz, R., 2017. \textit{An approximation of the backpropagation algorithm in a predictive coding network}. Nature Communications, 8(1), p.15610.

\bibitem{lecun2006tutorial}
LeCun, Y., Chopra, S., Hadsell, R., Ranzato, M. and Huang, F.J., 2006. \textit{A tutorial on energy-based learning}. In: Predicting Structured Data. MIT Press.

\bibitem{scellier2017equilibrium}
Scellier, B. and Bengio, Y., 2017. \textit{Equilibrium propagation: Bridging the gap between energy-based models and backpropagation}. Frontiers in Computational Neuroscience, 11, p.24.

\bibitem{yang2019physical}
Yang, G., 2019. \textit{Scaling limits of wide neural networks with weight sharing: Gaussian process behavior, gradient independence, and neural tangent kernel derivation}. arXiv preprint arXiv:1902.04760.

\bibitem{behrmann2019invertible}
Behrmann, J., Grathwohl, W., Chen, R.T.Q., Duvenaud, D. and Jacobsen, J.H., 2019. \textit{Invertible residual networks}. In: Proceedings of the 36th International Conference on Machine Learning (ICML), pp.573--582.

\bibitem{gallego2017neural}
Gallego, J.A. and Perich, M.A., 2017. \textit{Neural manifolds for the control of movement}. Neuron, 94(5), pp.978--984.

\bibitem{1}
Ye, M., Yang, D., Huang, Q., Kanski, M., Axel, L. and Metaxas, D.N., 2023. \textit{SequenceMorph: A unified unsupervised learning framework for motion tracking on cardiac image sequences}. IEEE Transactions on Pattern Analysis and Machine Intelligence, 45(8), pp.10409--10426.

\bibitem{2}
Delahunt, C.B., Riffell, J.A. and Kutz, J.N., 2018. \textit{Biological mechanisms for learning: a computational model of olfactory learning in the Manduca sexta moth, with applications to neural nets}. Frontiers in Computational Neuroscience, 12, p.102.


\bibitem{3}
Chen, J.R., 1991. \textit{Theory and applications of artificial neural networks}, Doctoral dissertation, Durham University.

bibitem{delahunt2017}
Delahunt, C.B., 2017. \textit{Smart as a Bug: A Computational Model of Learning in the Moth Olfactory Network, with Applications to Neural Nets}. Doctoral dissertation.

\bibitem{shavlik1994}
Shavlik, J.W., 1994. \textit{Combining symbolic and neural learning}. \textit{Machine Learning}, 14, pp.321--331.

\bibitem{chavlis2025}
Chavlis, S. and Poirazi, P., 2025. \textit{Dendrites endow artificial neural networks with accurate, robust and parameter-efficient learning}. \textit{Nature Communications}, 16(1), p.943.

\bibitem{zhuang2025}
Zhuang, H., Lin, Z., Yang, Y. and Toh, K.A., 2025. \textit{An analytic formulation of convolutional neural network learning for pattern recognition}. \textit{Information Sciences}, 686, p.121317.



\bibitem{shen2025}
Shen, T., Wang, J. and Zhang, X., 2025. \textit{Knowledge distillation via adaptive meta-learning for graph neural network}. \textit{Information Sciences}, 689, p.121505.


\bibitem{terres2024ecai}
Terres-Escudero, E.B., Del Ser, J. and Garcia-Bringas, P., 2024. \textit{On the improvement of generalization and stability of forward-only learning via neural polarization}. In: \textit{ECAI 2024}, pp.1919--1926. IOS Press.

\bibitem{zhuang2024f}
Zhuang, H., Liu, Y., He, R., Tong, K., Zeng, Z., Chen, C., Wang, Y. and Chau, L.P., 2024. \textit{F-OAL: Forward-only online analytic learning with fast training and low memory footprint in class incremental learning}. \textit{Advances in Neural Information Processing Systems}, 37, pp.41517--41538.



\end{thebibliography}

\end{document}